# Sistema de Gestión de Demanda y Consumo de Energía Eléctrica

[1]**Juan Ojeda Sarmiento**, [2]**Franco Pajares**
Departamento de Mejora Continua de Southern Copper Corporation
Av. Caminos del Inca N° 171, Santiago de Surco, Lima, Perú, 2008

**Resumen**
Este trabajo describe el sistema de gestión de demanda y consumo de energía eléctrica y su aplicación en la fundición de Southern Peru, que está compuesto por un modulo predictor de demanda eléctrica y un módulo de simulación del sistema eléctrico de la planta. El primero se construyó utilizando redes neuronales recurrentes, con algoritmos de aprendizaje backpropagation, capaz de pronosticar la evolución horaria de la demanda eléctrica con un porcentaje de error cercano al 1%. Esta información permite gestionar los picos de demanda antes de que estos se presenten, para su distribución tentativa en otros horarios o mejorar la tecnología de aquellos equipos que levantan la carga eléctrica. El módulo de simulación está basado en técnicas de estimación paramétrica, redes neuronales y ecuaciones matemáticas obtenidas por regresión estadística, que simula el comportamiento del sistema de energía eléctrica de la fundición. Estos módulos facilitan una adecuada planificación de la demanda y el consumo, pues permiten conocer el comportamiento de la demanda horaria y los patrones de consumo de la planta, incluyendo los componentes de la facturación; pero además las deficiencias energéticas y oportunidades de mejora, sobre la base del análisis de la información acerca de los equipos, procesos y planes de producción, así como programas de mantenimiento. Finalmente se presentan los resultados de su aplicación en la fundición de Southern Peru.

**Abstract**
This project describes the electricity demand and energy consumption management system and its application to Southern Peru smelter. It is composed of an hourly demand-forecasting module and of a simulation component for a plant electrical system. The first module was done using dynamic neural networks with backpropagation training algorithm; it is used to predict the electric power demanded every hour, with an error percentage below of 1%. This information allows efficient management of energy peak demands before this happen, distributing the raise of electric load to other hours or improving those equipments that increase the demand. The simulation module is based in advanced estimation techniques, such as: parametric estimation, neural network modeling, statistic regression and previously developed models, which simulates the electric behavior of the smelter plant. These modules facilitate electricity demand and consumption proper planning, because they allow knowing the behavior of the hourly demand and the consumption patterns of the plant, including the bill components, but also energy deficiencies and opportunities for improvement, based on analysis of information about equipments, processes and production plans, as well as maintenance programs. Finally the results of its application in Southern Peru smelter are presented.

---

[1] Industrial Engineer at *University of Lima* (Perú), M.Sc. in Systems Engineering at *National University of Engineering* (Perú), M.Sc. in Energy-management at *University of Koblenz-Landau (Germany)*, doctorate in Control, Artificial Vision and Robotic at *Polytechnic University of Catalonia* (Barcelona-Spain). Community Relations Superintendent of Southern Copper Corporation.
[2] Mechatronic Engineer at *National University of Engineering*, Lima-Perú, Project Engineer of Southern Copper.

## I. Introducción:

La energía eléctrica es un recurso estratégico en la producción industrial, su uso se extiende a todas las ramas de la actividad económica, y es un componente importante de la estructura de costos de las empresas industriales.

En la facturación, se consideran generalmente tres conceptos de pago, que son los siguientes:

**a)** Demanda (kW), que representa el cargo por la cantidad total de energía activa que se requiere al mismo tiempo de la red. Su composición depende del resultado de la formula llamada "demanda facturable" multiplicado por el costo del "kW" en el mes corriente.

La demanda facturable está influida por la máxima demanda del período, que se calcula tomando el mayor valor en un espacio de 15 minutos, medidos cada cinco minutos. Normalmente presenta oscilaciones horarias, diarias y estaciónales, estando marcada por la actividad productiva de la planta, de acuerdo al ejemplo de la figura N° 01.

Figura N° 01
**Diagrama de la carga diaria**

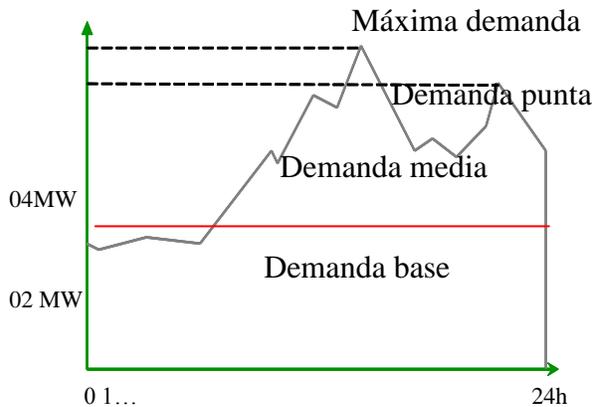

Fuente y elaboración: Brinkmann K., "Integration und Management dezentraler Energiversorgung"

**b)** Consumo (KWh), que se obtiene al multiplicar la potencia (kW) por el tiempo (h). También se obtiene al integrar el área debajo de la curva de la demanda.

**c)** Factor de potencia (%), que resulta la relación entre la potencia activa y la potencia aparente, siendo un número adimensional, comprendido entre 0 y 1.

Las empresas industriales en su mayor parte presentan sobrecostos por cargos de energía eléctrica, que se agravan por el aumento de los precios de energéticos convencionales en los mercados, generando crecientes montos de facturación mensual.

Para optimizar el uso de energía eléctrica, sobre la base de una adecuada planificación, es importante conocer los componentes de la facturación, el comportamiento de la demanda horaria y los patrones de consumo de la planta; pero además las deficiencias energéticas y oportunidades de mejora, que se obtiene recabando y analizando la información acerca de los equipos, procesos y planes de producción, así como programas de mantenimiento.

## II. Sistema de gestión de demanda y consumo de energía eléctrica:

El sistema de gestión de demanda y consumo de energía eléctrica se diseñó sobre la base de los requerimientos descritos en el apartado anterior; es decir, que incorpora la estructura tarifaria, brinda el pronóstico de la demanda horaria futura, y permite determinar los patrones de consumo, así como deficiencias energéticas de los centros de producción, facilitando ensayar oportunidades de ahorro para hacer un uso eficiente de la energía eléctrica.

Está compuesto por un modulo de previsión de demanda y otro de simulación de consumo de energía eléctrica (ver figura N° 02) que incorporan datos de la estructura tarifaria.

Figura N° 02
**Modelo de aplicación**

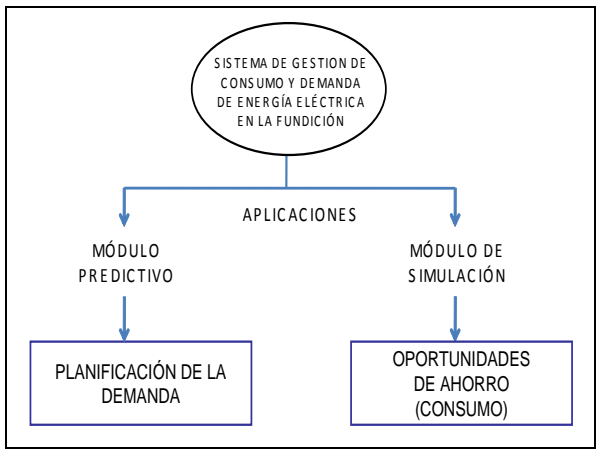

Fuente y elaboración propias

El módulo predictivo de la demanda de energía eléctrica horaria está construido con redes neuronales artificiales y permite predecir el comportamiento de la demanda horaria futura con un margen de error de 1 %, habiendo demostrado su superioridad sobre los procedimientos convencionales de regresión lineal y el "support vector machines".

Conocer la demanda futura de estos centros de costos significa disponer de una información valiosa, desde un punto de vista operativo, pero también desde un punto de vista estratégico, pues facilitar la gestión de los picos de demanda antes de que estos se presenten, permitiendo su distribución planificada en otros horarios o mejorar la tecnología de aquellos equipos que levantan la carga eléctrica [1].

El módulo de simulación de consumo de energía eléctrica hace posible conocer los patrones de consumo, identificando deficiencias energéticas, para determinar oportunidades de ahorro y ensayar modificaciones a futuro en el sistema, como: cambios, repotenciación o mejora de performance de equipos, sin necesidad de llevarlos a la práctica, elevando de esta forma las probabilidades de éxito de cualquier propuesta de mejora.

## 2.1 Desarrollo del Módulo predictor:

Como antecedentes de esta aplicación, se pueden citar los siguientes trabajos:

La investigación titulada "Time series forecasting: Obtaining long term trends with self-organizing maps", es decir, predicción de series temporales para obtener tendencias a largo plazo, basado en mapas Kohonen de auto-organización [2].

Igualmente importante resultan los trabajos de modelos predictivos para prevenir la intensidad de los huracanes, elaborado por Venero Castro [3].

Continúa la lista con el trabajo de Edgar Sánchez, Alma Alanis y Jesús Rico, denominado "*Predicción de la demanda eléctrica, usando redes neuronales entrenadas por filtro de Kalman*", En dicha investigación, se desarrolló un modelo predictivo basado en redes neuronales artificiales, tipo perceptrón multicapa recurrente, con el uso del filtro de Kalman Extendido (EKF) para el entrenamiento de una red MLP (multilayer perceptron)[4] y tecnologías de apoyo a la administración del territorio, desarrollado por Simeao de Medieros [5].

El trabajo de Pérez Ortiz [6] para modelos predictivos, basados en redes neuronales recurrentes de tiempo discreto. Asimismo, igualmente importantes para este trabajo fueron las investigaciones de Hufendiek Kai [7] y Meisenbach Christ [8], sobre aplicación de redes neuronales a la predicción de carga eléctrica, elaborado para "Institut für Energiewirtschaft und Rationelle Energieanwendung", y la investigación de Petridis and A. Kehagias [*9*] sobre series temporales.

Como caso de aplicación, se tomó la planta de Fundición de SOUTHERN PERU [*10*], pues las variables a considerar en este caso se asemejan a las de cualquier planta industrial del país.

Para el diseño de la red neuronal dinámica recurrente (con feedback) NARX modificada, se consideraron los siguientes componentes:

## 2.1.1 Definición de patrones:

Los patrones o entradas a utilizar son:

**Calendario.-** Será un array de vectores de 5 x N elementos, donde 5 son el número de patrones calendario y N es el número de muestras por mes, teniendo como periodo de muestreo 15 minutos: N = 2976 (31 días) **y** N = 2880 (30 días). Los patrones considerados en este caso fueron: cuarto de hora, hora, día, semana y mes.

**Producción.-** Será un vector de 3 x M, donde 3 son el número de patrones y M es el número de muestras, teniendo como periodo de muestreo 1 día.

**Demanda del mes anterior.-** Será un vector de 1 x N, **donde** 1 es el único patrón y N es el número de muestras.

Teniendo como base los patrones descritos anteriormente y revisando pequeñas muestras de data histórica de la demanda eléctrica en la Fundición Ilo, se procedió a evaluar la influencia de los distintos patrones, sobre la base de literatura especializada [*11-14*]:

Figura N° 03
**Diagrama de bloques módulo predictivo**

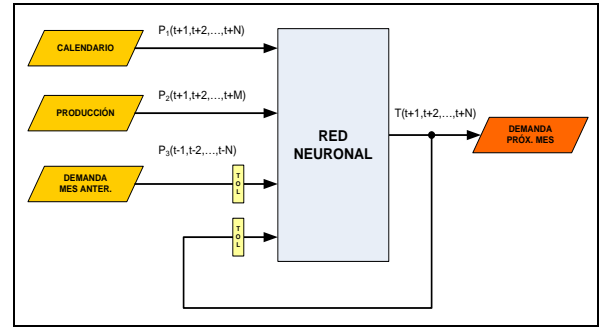

Fuente y elaboración propias

### a) Calendario:

#### a.1) Cuarto de hora:
Este patrón se consideró debido a que la facturación se realiza teniendo como base los cuartos de hora. Para tal fin se tomó una muestra de 1344 datos entre el lunes 18 de junio y el domingo 01 de julio del 2007 (96 cuartos de hora por día).

#### a.2) Hora del día:
La hora del día influye en la curva de demanda eléctrica de una ciudad, debido a que en horas punta, como las 3 de la tarde, por ejemplo, el consumo es mayor; sin embargo en el medio industrial la demanda eléctrica no sufre cambios significativos. En este caso, se tomó cada hora entre el 18 de junio y el 24 de junio del 2007

#### a.3) Día de la Semana:
El día de la semana influye de gran manera en el comportamiento de la demanda, ya que si es un día programado para colada, y a su vez todos los demás procesos no dejan de funcionar, la demanda media y máxima se verán incrementadas.

Para este estudio, se tomaron la demanda media y máxima diaria durante 10 semanas, comprendidas entre el 23 de abril al 01 de julio del 2007, cuyos resultados se presentan gráficamente entre las figuras N° 04 a N° 07.

Figura N° 04
**Demanda media semanal (23/04-01/07)**

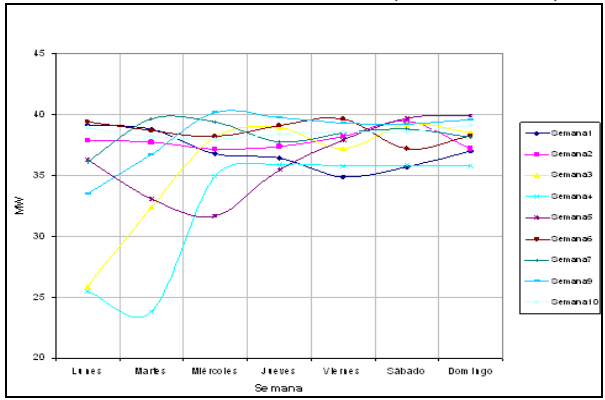

Fuente: Servidor IION
Elaboración: Propia

Figura N° 05
**Demanda máxima semanal (23/04-01/07/07)**

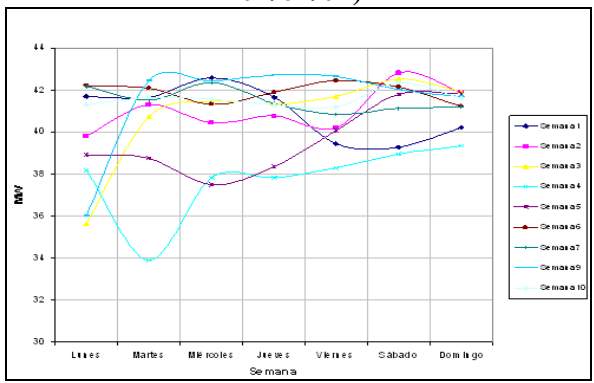

Fuente: Servidor IION
Elaboración: Propia

Figura N° 06
**Demanda media diaria (23/04-01/07/07)**

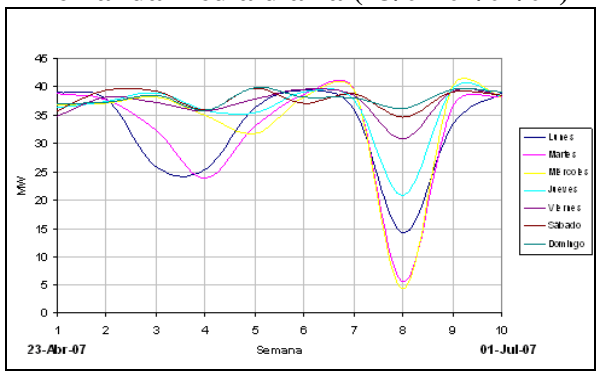

Fuente: Servidor IION
Elaboración: Propia

Figura N° 07
**Demanda media diaria en 10 semanas de 23 de abril al 01 de julio del 2007**

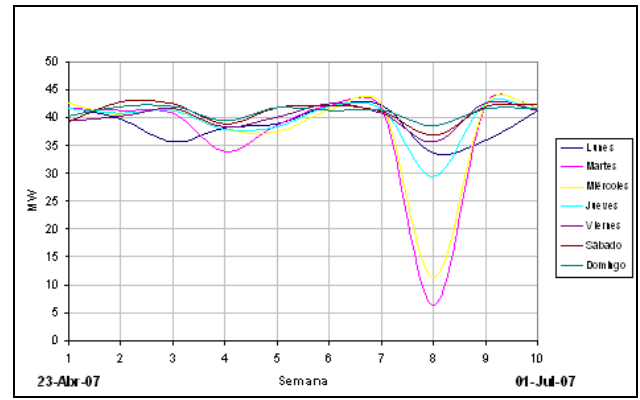

Fuente: Servidor IION
Elaboración: Propia

En las figuras N° 04 y 05, no se observa ninguna relación precisa tanto en la demanda media semanal, así como en la demanda máxima semanal, por lo que se concluye, que los días - vistos de manera continua por semana - no presentan gran influencia en la demanda; sin embargo, si se aprecian las semanas de manera continua, como en los figuras N° 06 y N° 07, se observa, sobre todo en el caso de la demanda máxima, que dependiendo de los días, la curva sigue cierta periodicidad, convirtiendo a este factor en uno de los más importantes para el análisis.

**b)    Producción:**
La producción es un factor imprescindible en el medio industrial, pues - dependiendo de ésta y sus estimados - se presupuestan los cargos por concepto de insumos, combustibles, electricidad y demás. Por lo tanto éste es un parámetro principal en nuestro análisis.

**b.1)    Producción de ánodos:**
Para este estudio se realizó un análisis del la producción diaria, versus la media y la demanda máxima día a día entre el 1ro de marzo al 30 de junio del 2007. Los resultados fueron los siguientes:

Figura N° 08
**Producción ánodos vs. demanda media**

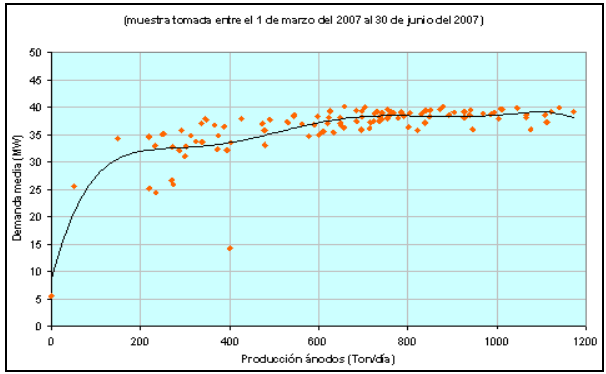

**Fuente:** Reporte de producción planta de ánodos.
**Elaboración:** Propia

Figura N° 09
**Producción ánodos vs. demanda máxima**

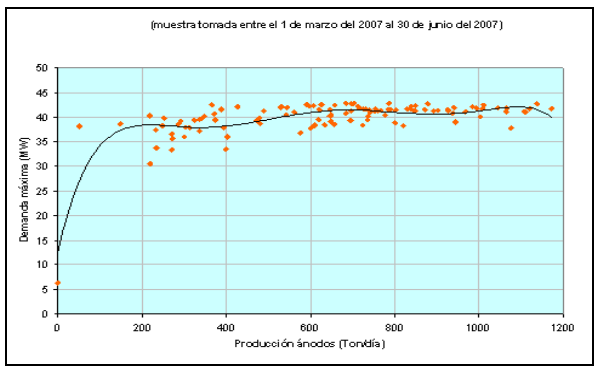

Fuente: Reporte de producción planta de ánodos.
Elaboración: Propia

En la figura N° 08, se observa que la curva de demanda media diaria varía de manera proporcional a la producción diaria de ánodos de cobre, siguiendo una tendencia cuasi logarítmica.

A su vez, si se revisa la figura N° 09, se aprecia como la demanda máxima diaria sigue casi la misma tendencia que la demanda media. Por lo tanto, se concluye que este factor es determinante en la curva de demanda mensual. Otro punto a que hace necesario su elección es el hecho de conocer los valores estimados o requeridos de producción en el futuro, hacen de éste un patrón indispensable en el modelo predictivo.

**b.2) Producción de ácido sulfúrico:**
Debido a ser otro de los productos de la fundición y ser el resultados de 2 de las plantas de mayor consumo eléctrico (Plantas de Ácido 1 y 2).

Figura N° 10
**Producción de ácido sulfúrico vs. demanda media y máxima de fundición de Ilo**

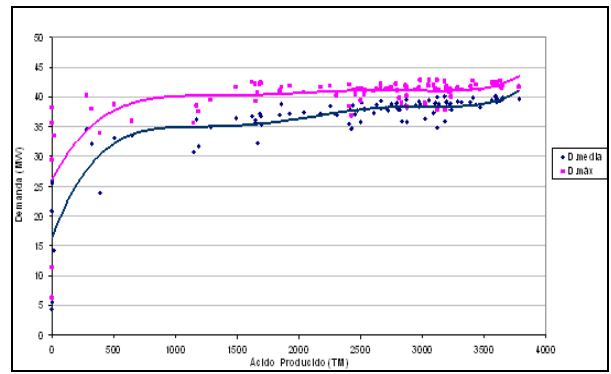

Fuente: Reporte de producción planta de ácido
Elaboración: Propia

**b.3) Producción de oxígeno:**
Debido al gran consumo de las plantas de oxígeno 1 y 2, su influencia en producción se vuelve imprescindible, para nuestro modelo.

Figura N° 11
**Producción oxígeno vs. demanda media y máxima de la Fundición**

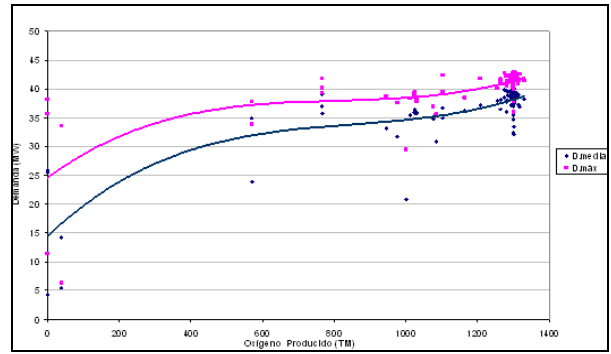

Fuente: Reporte de producción planta de ácido
Elaboración: Propia

**c) Demanda anterior:**

En el campo del planeamiento de presupuestos se realizan estimaciones futuras, basándose en los últimos valores obtenidos. A su vez el modelo a desarrollar, también considerará la demanda eléctrica del último mes. A continuación se evaluará la influencia de la demanda anterior en la demanda presente de los meses entre marzo y junio del 2007.

Figura N° 12
**Influencia de demanda media diaria de marzo en demanda media diaria de abril**

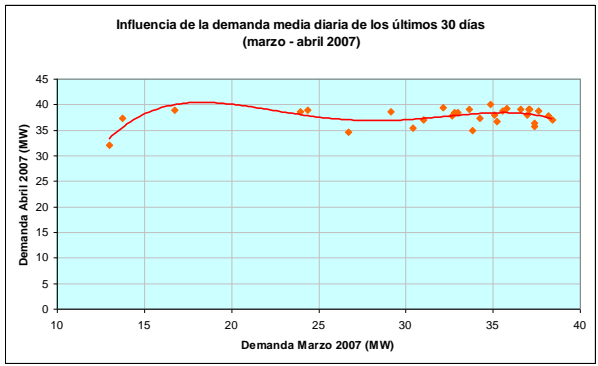

Fuente: Reporte de producción planta de ácido
Elaboración: Propia

Figura N° 13
**Influencia de la demanda media diaria de abril en demanda media diaria de mayo**

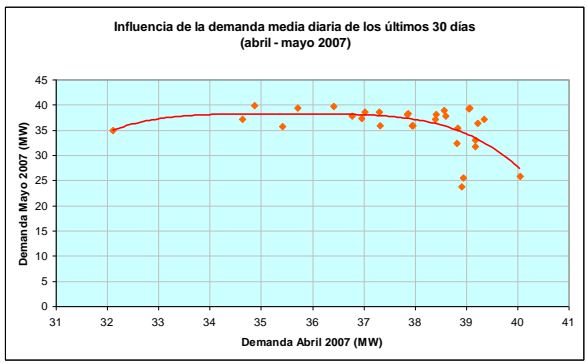

Fuente: Servidor ION
Elaboración: Propia

Figura N° 14
**Influencia de demanda media diaria de mayo en demanda media diaria de junio**

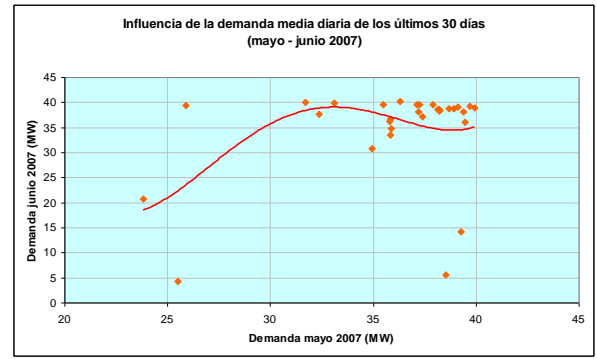

**Fuente:** Servidor ION
**Elaboración:** Propia

### 2.1.2 Selección de factores para el modelo predictivo:

Luego del análisis realizado en el punto 2, se concluye que existen factores o patrones que infieren de manera predominante que los demás. Posteriormente al estudio de las figuras del 2 al 14 y sus resultados, se obtiene la siguiente tabla de influencia:

Tabla N° 01
**Tabla de patrones seleccionados
en curva de demanda de energía eléctrica**

|  | Influencia | | Nivel de influencia |
|---|---|---|---|
|  | Linealidad | Proporcion. |  |
| 1. Calendario | No | No | Medio |
| 1.1. Mes | No | No | Bajo |
| 1.2. Semana | No | No | Medio |
| 1.3. Día | No | No | Medio |
| 1.4. Hora | No | No | Alto |
| 1.5. Cuarto | No | No | Medio |
| 2. Producción | No | Sí | Muy alto |
| 2.1. Ánodos de cobre | No | Sí | Muy alto |
| 2.2. Ácido sulfúrico | No | Sí | Alto |
| 2.3. Oxígeno | No | Sí | Alto |
| 3. Demanda anterior | No | Sí | Medio |

Fuente: Fundición de Southern Peru
Elaboración: propia

Como se ve los factores que influyen directamente en la demanda eléctrica son los relacionados a la producción, calendario,

demanda anterior y temperatura; sin embargo, debido a la dificultad de tener valores exactos de la temperatura en el futuro, sólo se consideran los tres primeros.

Los factores considerados en la RNA del modelo de predicción se muestran en la tabla N° 02.

Tabla N° 02
**Factores de la red neuronal**

| Patrón | Unidades |
|---|---|
| **1. Calendario** | |
| 1.1. Mes | 0 - 11 |
| 1.2. Semana | 0 - 4 |
| 1.3. Día | 0 - 6 |
| 1.4. Hora | 0 - 23 |
| 1.5. Cuarto de hora | 0 - 3 |
| **2. Producción** | |
| 2.1. Ánodos | TM/h |
| 2.2. Ácido | TM/h |
| 2.3. Oxígeno | TM/h |
| **3. Demanda anterior** | |
| 3.1. Demanda mes anterior | MW |

Fuente y elaboración propias

### 2.1.3   Diseño de la RNA:

Una vez definida el tipo de RNA a utilizar (dinámica recurrente), así como sus patrones y parámetros, procedemos a diseñar la nueva arquitectura de la RNA. A continuación se muestra el diagrama detallado de esta arquitectura, considerando todos los enlaces y realimentaciones.

Figura N° 15
**Arquitectura final de la RNA**

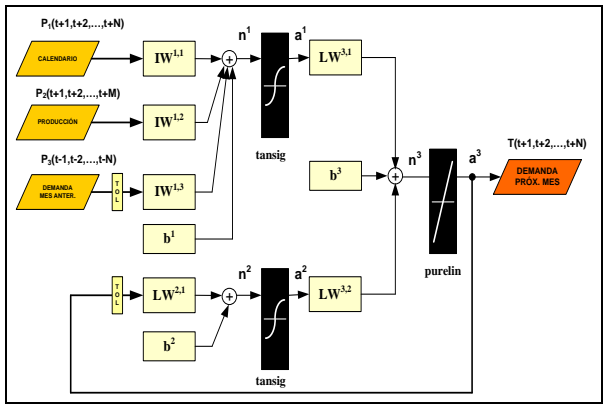

Fuente y elaboración propias

### 2.1.4   Interfaz gráfica:

Figura N° 16
**Módulo predictor de demanda eléctrica**

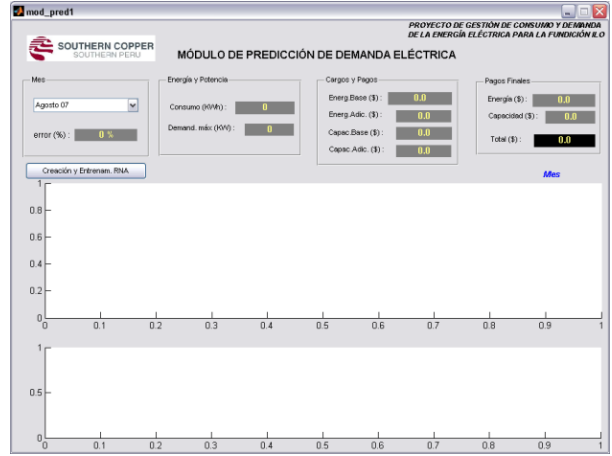

Fuente y elaboración propias

### 2.1.5   Validación:

En la figura 5 se muestra el comportamiento de la demanda predicha para el mes de octubre 2007.

Figura N° 17
**Módulo predicción de demanda eléctrica**

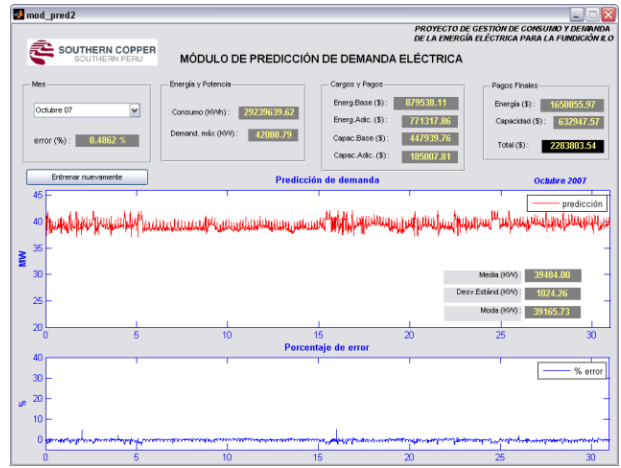

Fuente y elaboración propias

La data obtenida para la validación es la data oficial del servidor IION, y que consta en los reportes del departamento de Sistemas de Potencia.

Para la validación del módulo se utilizó el error cuadrático medio. El error cuadrático medio en porcentaje se define como:

$$\%e_{rms} = \frac{\sum_{i=1}^{N}\left(\frac{p_i - r_i}{r_i}\right)^2}{N} \times 100 \quad (1)$$

Donde:
$p_i$: Dato simulado
$r_i$: Dato real
N: Número de muestras

La tabla 3 muestra el error cuadrático medio de toda la serie de datos de octubre de 2007(0,4862 %).

Tabla N° 03
Error cuadrático medio

| Mes | %erms |
|---|---|
| Oct-07 | 0.4862% |

Este indicador representa un error de 1 % respecto de la demanda y 3% con relación a la energía consumida, comparándola con la facturación real de dicho mes, como se aprecia en las tablas N° 04.

Tabla N° 04
Error de aproximación

| Mes | $E_{predicha}$ (KWh) | $E_{real}$ (KWh) | %error |
|---|---|---|---|
| Oct-07 | 29,239,640 | 28,336,177 | 3.09% |

| Mes | $D_{predicha}$ (KW) | $D_{real}$ (KW) | %error |
|---|---|---|---|
| Oct-07 | 42,089 | 42,569 | -1.14% |

## 2.2 Desarrollo del módulo de simulación:

El módulo de simulación se construye para cada planta, teniendo en cuenta relaciones mecánicas, eléctricas y termodinámicas, incluyendo el componente productivo existente, sobre la base de ecuaciones matemáticas.

Este aplicativo es una herramienta de gestión necesaria para conocer los patrones de consumo de las plantas, identificando deficiencias energéticas, que permitan encontrar oportunidades de ahorro y ensayar propuestas de mejora.

Este módulo de simulación replica los procesos y subprocesos energéticos, desde el punto de vista energético, pero además incluye componentes productivos, considerando las siguientes ecuaciones físicas:

Ejemplo de circuitos eléctricos:

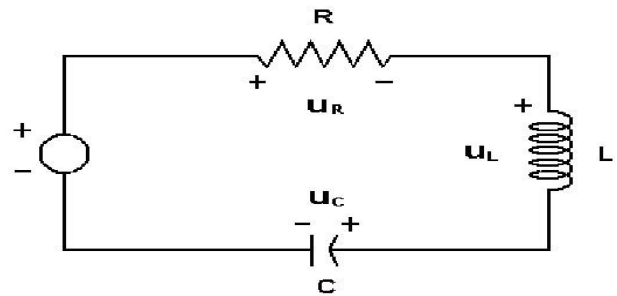

Fórmulas básicas:

$$u_R = i \cdot R$$
$$u_L = L \cdot \frac{di}{dt}$$
$$i = C \cdot \frac{du_C}{dt}$$
$$u_C = \frac{1}{C} \cdot \int i \cdot dt$$

$x_1 = i$
$x_2 = u_c$

Leyes de Kirchhoff
$$-u + u_K + u_L + u_C = 0$$

$$u = R_i + L\frac{di}{dt} + u_C$$

$$i = C \cdot \frac{du_C}{dt}$$

Entrada: $u(t)$    Salida: $i(t)$

Ecuaciones de estado:

$$\dot{x} = f(x,u)$$

De las entradas:

Primera ecuación de estado

$$\dot{x}_1 = \frac{1}{L}(u - Rx_1 - x_2)$$

Segunda ecuación de estado:

$$\dot{x}_2 = \frac{1}{C} \cdot x_1$$

**Ecuaciones de estado de la salida:**

$$y = h(x,u) = x_1$$

**Ecuaciones de estado en forma matricial:**

$$\begin{bmatrix} \dot{x}_1 \\ \dot{x}_2 \end{bmatrix} = \begin{bmatrix} -R/L & -1/L \\ 1/L & 0 \end{bmatrix} \cdot \begin{bmatrix} x_1 \\ x_2 \end{bmatrix} + \begin{bmatrix} 1/L \\ 0 \end{bmatrix} \cdot u$$

Ecuación de salida:

$$y = \begin{bmatrix} 1 & 0 \end{bmatrix} \cdot \begin{bmatrix} x_1 \\ x_2 \end{bmatrix} + [\;] \cdot u$$

Ecuaciones diferenciales $\rightarrow$ Ecuaciones de estado

a) $\dfrac{d^n y}{dt^n} = y^{(n)} = h(y, \dot{y}, \ddot{y}, ..., y^{(n-1)}, u)$

$\rightarrow$ Ecuación diferencial de orden $n$

$$\left. \begin{aligned} x_1 &= y \\ x_2 &= \dot{y} \\ x_3 &= \ddot{y} \\ &\vdots \\ x_n &= y^{(n-1)} \end{aligned} \right\} \Rightarrow \begin{aligned} \dot{x}_1 &= x_2 \\ \dot{x}_2 &= x_3 \\ &\vdots \\ \dot{x}_{(n-1)} &= x_n \\ \dot{x}_n &= h(x_1, x_2, ..., x_n, u) \end{aligned}$$

Salida:  $y = x_1$

b) Dos ecuaciones diferenciales de segundo orden:

$$\ddot{y}_1 = h_1(y_1, \dot{y}_1, y_2, \dot{y}_2, u)$$
$$\ddot{y}_2 = h_2(y_1, \dot{y}_1, y_2, \dot{y}_2, u)$$

Variables de estado:

$$\left. \begin{aligned} x_1 &= y_1 \\ x_2 &= \dot{y}_1 \\ x_3 &= y_2 \\ x_4 &= \dot{y}_2 \end{aligned} \right\} \Rightarrow \begin{aligned} \dot{x}_1 &= x_2 \\ \dot{x}_2 &= h_1(x_1, ..., x_4, u) \\ \dot{x}_3 &= x_4 \\ \dot{x}_4 &= h_2(x_1, ..., x_4, u) \end{aligned}$$

Ecuaciones de salida: $y_1 = x_1$, $y_2 = y_3$

Este módulo se aplicó inicialmente en la fundición de Southern Peru, con los resultados que se presentan a continuación. En la figura N° 28, se muestra el sistema de energía eléctrico de la Fundición y sus plantas.

**Figura N° 18**
**Diagrama de bloques del módulo de simulación (Southern Peru)**

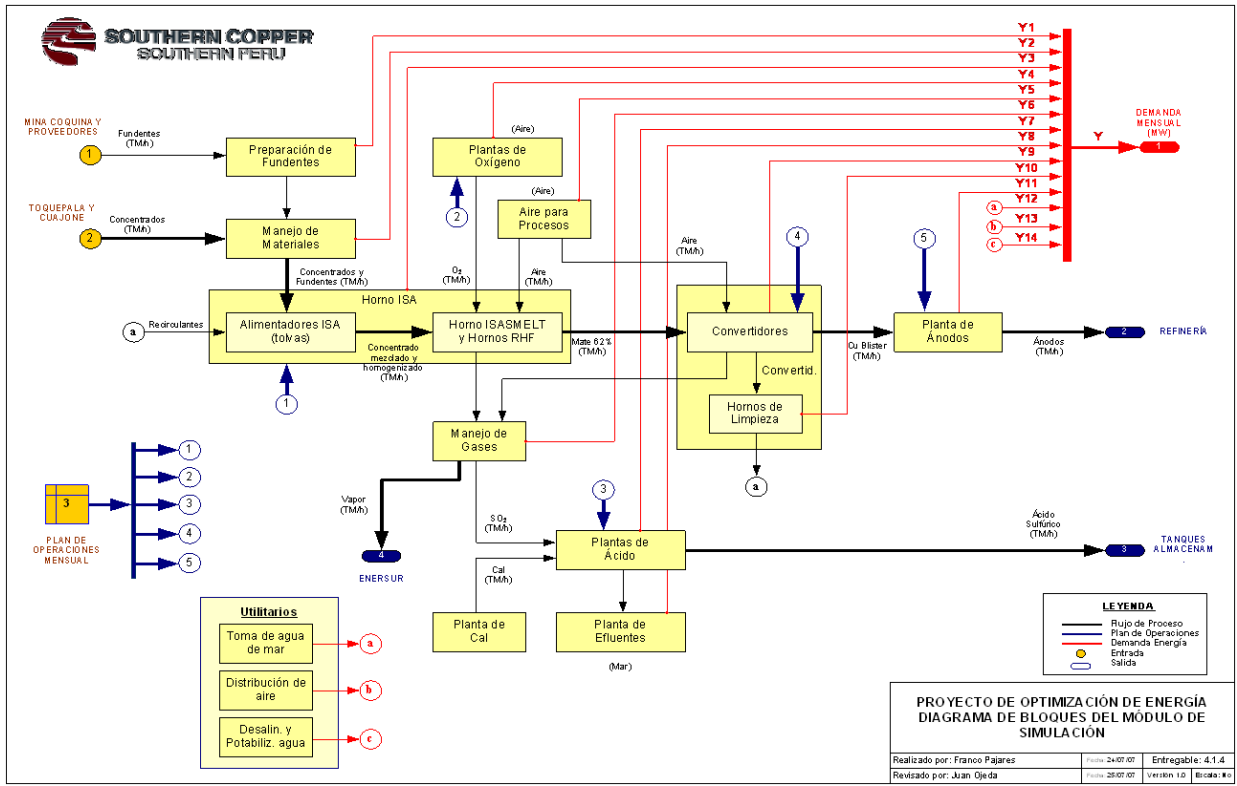

Fuente: Southern Peru
Elaboración: propia

A partir de la simulación de la planta de fundición fue posible conocer los patrones de consumo de cada centro de costos, incluyendo sus deficiencias energéticas y oportunidades de ahorro [15].

El modelamiento determinó las siguientes deficiencias:

- Compresoras funcionando fuera del punto de operación recomendado.
- Deficiente regulación de flujo en Blowers.
- Mala regulación de la temperatura de baño del horno de fusión de concentrados de tecnología Isasmelt.
- Sobredimensionamiento de bombas de en la planta de Toma de Agua de Mar.

Este diagnóstico permitió, asimismo, brindar alternativas de solución en cada caso, que se resumen en las siguientes propuestas

**2.2.1 Plantas de oxígeno 1 y 2 (43.35% del consumo):**

a) Seteo y ajuste de las válvulas IGV de las compresoras MAC a condiciones de diseño; es decir, corregir los lazos de control de la válvula de ingreso de alabes (IGV) de las compresoras MAC, para llevar el flujo y presión de descarga del aire a las condiciones de diseño, de tal manera que se aumente la eficiencia eléctrica del motor, disminuyendo las pérdidas mecánicas por fricción y calor.

Esta modificación genera una disminución de la demanda de 91.75 KW por el compresor A

y 282.83 KW por el compresor B, bajando el consumo de energía en 269,697 KWh.

Figura N° 19
**Condiciones actual y de diseño MAC**

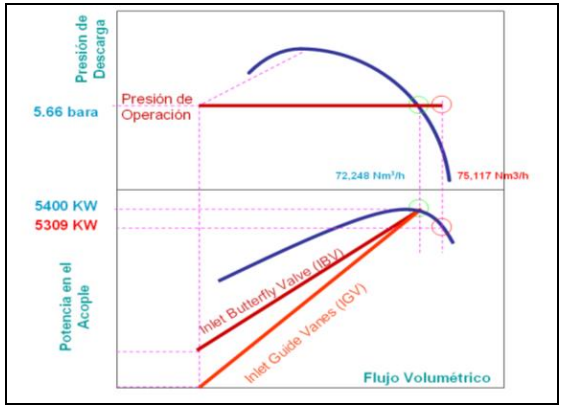

**Fuente:** Curvas estándares para compresores Ingesroll
**Elaboración:** Propia

b) Regulación de flujo por variadores de velocidad en las compresoras MAC; es decir, cambiar el método de regulación de flujo de aire en las compresoras MAC K111 A/B, pasando del uso de las válvulas de ingreso de alabes (IGV) a la utilización de variadores de velocidad en los motores de inducción D111 A/B.

Figura Nº 20
**Balance de Oxígeno de la Fundición Ilo**

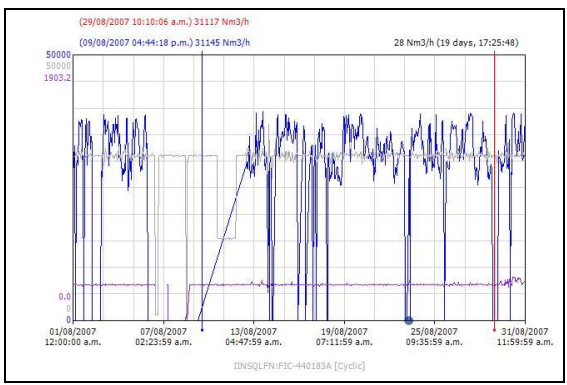

**Fuente:** Active Factory
**Elaboración:** Propias

Figura Nº 21
**Lazo de control de las MAC utilizando variadores de velocidad**

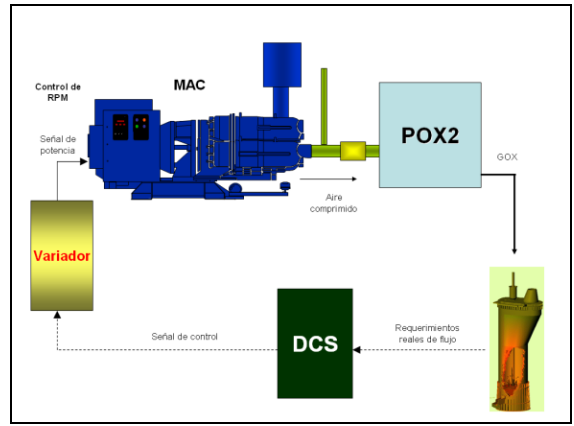

Fuente y elaboración propias

Esta modificación genera una reducción de demanda promedio: de 3,144 KW. Y una disminución de consumo mensual de 2, 263, 680 KWh.

### 2.2.2 Plantas de ácido 1 y 2 (23.10% del consumo):

Regulación de flujo por variadores de velocidad en los Blowers A y B, mediante variadores de frecuencia para los Blowers A y B de la planta de ácido N° 2, regulando el flujo de gases succionados en función de los flujos provenientes de la Fundición (Horno ISA y Convertidores).

Figura N° 22
**Regulación de flujo por variadores de velocidad en los Blowers**

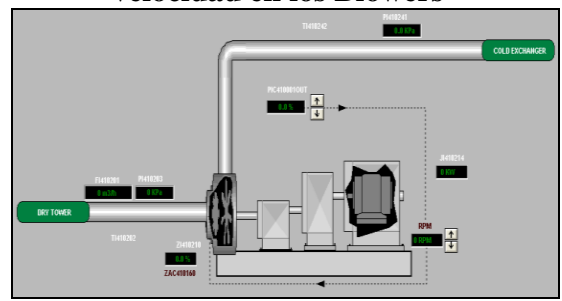

**Fuente y elaboración propias**

Esta mejora proporciona una reducción de demanda promedio: 1,416 KW y una disminución de consumo mensual de 1, 019, 520 KWh.

### 2.2.4 Toma de agua de mar (5.42% del consumo):

Actualmente se usan 2 bombas de 1250 HP que permite disponer de un flujo en exceso innecesario para elevar el agua hasta todos los sistemas de refrigeración con una operación. Por lo que se propone usar bombas de 625 HP para tener una mayor flexibilidad en la disponibilidad de caudal.

El encendido y apagado de estas bombas se realizaría según la demanda de agua de refrigeración en la planta.

Figura N° 23
**a) Encendido de bombas según necesidad de caudal.**

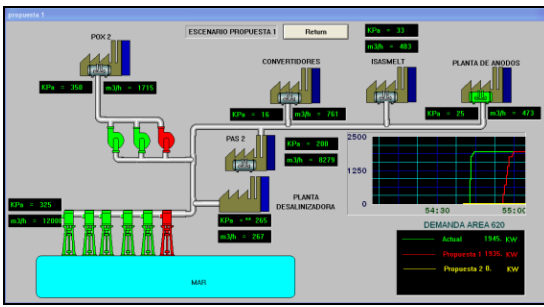

Fuente y elaboración propias

**b) Apagado de bombas según necesidad de caudal**

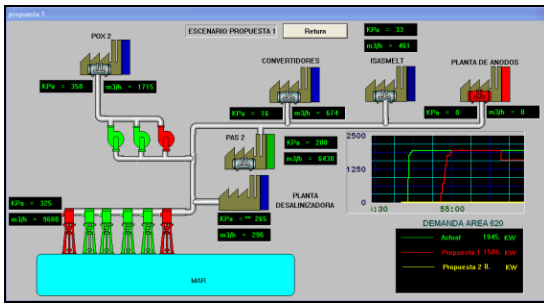

Fuente y elaboración propias

Al usar variadores de velocidad para regular el caudal necesario, se logra una mayor flexibilidad en variar el caudal proveído por el sistema de bombeo, lográndose una reducción de la demanda promedio de 359 KW y una disminución del consumo mensual de 129, 240 KWh

Figura N° 24
**Disminuir la demanda adicional**

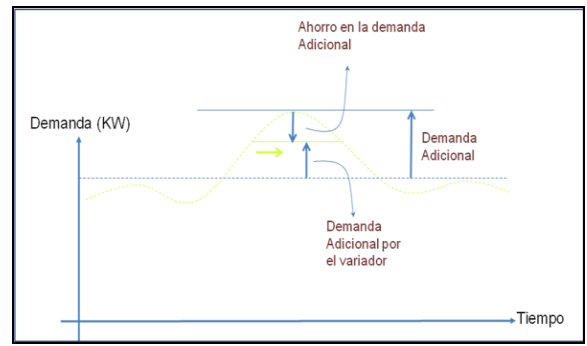

**Fuente y elaboración:** Propias

### 2.2.5 Horno Isasmelt (5.14% del consumo):

Para el horno, se planteó un controlador avanzado, mediante lógica difusa, que permitirá controlar eficazmente las oscilaciones bruscas de la temperatura, disminuyendo las paradas no programadas del sistema y eliminando sobrecostos de energía eléctrica, y ahorros por la menor reposición de refractarios en el horno.

### 2.3 Implementación de propuestas simuladas:

Todas estas propuestas, se incorporaron al módulo de simulación para conocer su repercusión en el proceso productivo de la fundición y validar los cálculos económicos de dichas mejoras, cuyos detalles se puede apreciar en el siguiente cuadro de ahorros mensuales estimados (ver tabla N° 05).

Tabla N° 05
**Ahorros estimados**

| N° | PLANTA | PROPUESTA | Reducción demanda promedio (KW) | Reducción Consumo mensual (KWH) | Ahorro por cargo de demanda adicional (7.5985$/KW) | Ahorro por cargo de energía adicional (0.074$/KWh) | AHORRO TOTAL US$ |
|---|---|---|---|---|---|---|---|
| 1 | POX2 | Seteo y ajuste de las válvulas IGV de las compresoras MAC a condiciones de diseño | 374 | 269,280 | 2,842 | 19,927 | 22,769 |
| 2 | POX2 | Regulación de flujo por variadores de velocidad en la compresora MAC K111 A | 1,599 | 1,151,280 | 12,150 | 85,195 | 97,345 |
| 3 | POX2 | Regulación de flujo por variadores de velocidad en la compresora MAC K111 B | 1,545 | 1,112,400 | 11,740 | 82,318 | 94,057 |
| 4 | PAS2 | Regulación de flujo por variadores de velocidad en el Blower A | 713 | 513,360 | 5,418 | 37,989 | 43,406 |
| 5 | PAS2 | Regulación de flujo por variadores de velocidad en el Blower B | 703 | 506,160 | 5,342 | 37,456 | 42,798 |
| 6 | PAS2 | Apagado automático de bombas, ventiladores y sopladores en días de producción nula | (166*) | 11,952 | 0 | 884 | 884 |
| 6 | Agua mar | Cambiar 3 bombas de agua de 933 KW por 6 bombas de 490 KW | 111 | 79,920 | 843 | 5,914 | 6,758 |
| | | **TOTAL MENSUAL** | 5,045 | 3,644,352 | 38,335 | 269,683 | 308,018 |

Fuente: Southern Peru
Elaboración: propias

Las propuestas de ahorro estimado suman aproximadamente US$ 308,018 mensuales, a un precio de 0.074 US$/KWh, sin considerar el ahorro de energía eléctrica por optimización del sistema de control del horno, que presenta ventajas por la mayor vida útil del refractario ($ 300,000 anuales) y la eliminación de sobrecostos anuales por la reducción de paradas no programadas:
Ahorro = factor * horas estimadas parada *

US$/KWh * KW= 0.8 * 229 h * 0.074 US$/KWh * 13 200 KW = **US$ 178,949.76**

### III. Conclusiones:

El rigor en el manejo de la información ha permitido el diseño, construcción y validación del sistema de gestión de demanda y consumo, con sus módulos predictivo y de simulación, facilitando la optimización de la demanda de potencia y consumo de energía eléctrica.

El modelo predictivo facilita la planificación de la demanda de energía eléctrica, por su potencia en el pronóstico de la evolución horaria de la energía eléctrica, haciendo posible, más adelante, implementar un control de demanda.

El módulo de simulación permitió integrar los subsistemas más importantes dentro del proceso de la Fundición, como son: el eléctrico, el mecánico, el de instrumentación y el energético, de tal manera que se pudieron identificar deficiencias energéticas (eléctricas) y ensayar propuestas de mejora en la operación y control, con el fin de evaluar su impacto en todos los subsistemas mencionados.

Fue posible identificar y validar oportunidades de ahorro de consumo de energía eléctrica de la Fundición de Ilo en aproximadamente 20 % de su facturación mensual.